\documentclass[nohyperref]{article}

\usepackage{microtype}
\usepackage{graphicx}
\usepackage{subfigure}
\usepackage{booktabs} 

\usepackage{hyperref}
\usepackage{xcolor}

\usepackage[accepted]{icml2022}

\usepackage{amsmath}
\usepackage{amssymb}
\usepackage{mathtools}
\usepackage{amsthm}

\usepackage[capitalize,noabbrev]{cleveref}

\theoremstyle{plain}

\theoremstyle{definition}

\theoremstyle{remark}

\usepackage[textsize=tiny]{todonotes}

\icmltitlerunning{MRCLens: an MRC Dataset Bias Detection Toolkit}

\begin{document}

\twocolumn[
\icmltitle{MRCLens: an MRC Dataset Bias Detection Toolkit}




\begin{icmlauthorlist}
\icmlauthor{Yifan Zhong}{cmu}
\icmlauthor{Haohan Wang}{cmu,uiuc}
\icmlauthor{Eric P. Xing}{cmu,mbzuai}
\end{icmlauthorlist}

\icmlaffiliation{cmu}{School of Computer Science, Carnegie Mellon University}
\icmlaffiliation{uiuc}{School of Information Sciences, University of Illinois Urbana-Champaign}
\icmlaffiliation{mbzuai}{Mohamed bin Zayed University of Artificial Intelligence}

\icmlcorrespondingauthor{Yifan Zhong}{zhongyifan.eva@gmail.com}

\icmlkeywords{Machine Learning, ICML}

\vskip 0.3in
]



\printAffiliationsAndNotice{}  

\begin{abstract}
Many recent neural models have shown remarkable empirical results in Machine Reading Comprehension, but evidence suggests sometimes the models take advantage of dataset biases to predict and fail to generalize on out-of-sample data. While many other approaches have been proposed to address this issue from the computation perspective such as new architectures or training procedures, we believe a method that allows researchers to discover biases, adjust the data or the models in an earlier stage will be beneficial. Thus, we introduce \textbf{MRCLens}, a toolkit which detects whether biases exist before users train the full model. For the convenience of introducing the toolkit, we also provide a categorization of common biases in MRC. 
\end{abstract}

\section{Introduction}
The ability of machines to read and comprehend texts is a critical skill in natural language processing. Recently sophisticated neural network models such as BiDAF \citep{seo2016bidirectional}, RNet \citep{wang2017r} and QANet \citep{yu2018qanet} have achieved remarkable accuracies on several benchmark datasets like SQuAD \citep{rajpurkar2016squad}. However, some popular datasets contain superficial patterns that can be exploited by models to make predictions without learning much about the contexts \citep{wang2021measure}. As a result, the models might fail to generalize to out-of-sample datasets \cite{yogatama2019learning, rimell2009unbounded, paperno2016lambada} or in adversarial settings \citep{jia2017adversarial, wallace2019universal}.

The community has approached the problem 
from the modelling perspective
\citep{ fisch-etal-2019-mrqa, takahashi-etal-2019-cler}. 
For example, 
a popular example
is to first train a bias-only model based, 
and then combine it with a full model to learn the additional information \citep{sugawara2018makes}. 
In addition, there are also diagnostic tools such as interactive frameworks \citep{lee2019qadiver} or attention matrix visualizer \citep{ruckle2017end,liu2018visual} to evaluate QA models. 
A common limitation of these approaches 
is we cannot discover the biases until the full models have been trained and evaluated, 
which posted a challenge for such analysis when computational resources are limited.

Our study contributes to existing work by introducing a toolkit \textbf{MRCLens} which detects bias in MRC datasets. This toolkit tests a given dataset against several known biases before training the full model. It includes a relatively smaller model and thus can be applied when computational resources are limited. 
Our toolkit can be applied to various SQuAD formatted MRC datasets. This also allows researchers to make adjustments to improve the datasets or develop models that target the existing biases. 
Along our implementation of the toolkit, we find it convenient to categorize the biases. Thus, 
our second contribution is a summary of common biases in MRC. Through literature reviews, we identify various recurring biases which can fall into three categories. We summarize them as \textbf{Similarity Bias}, \textbf{Keyword Bias} and \textbf{Question Bias}. Furthermore, we introduce the concept of ‘distance’ as a way to measure MRC bias. These concepts will be discussed in more detail in section 2.2. 

\section{Background} 
\subsection{Related Work}
The MRC task evaluates a system’s ability to retrieve information and make meaningful inferences \citep{sutcliffe2013overview}. Many recent neural models have shown remarkable results, but some models exploit dataset-specific patterns which fail to generalize \citep{clark2019don, talmor2019multiqa, sen2020models}. Min \textit{et al}. observed that 92\% of answerable questions in SQuAD can be answered only using a single context sentence \citep{min2018efficient}. When confounding sentences which have semantic overlap with the question were added to a dataset, the MRC model’s performance dropped significantly \citep{jia2017adversarial}. In another experiment, many questions in an easier subset of the dataset had their answers in the most similar sentence and could be answered with word-matching \citep{sugawara2018makes}. In Story Cloze Test tasks, recognizing the superficial features is essential for the models to achieve good performance \citep{schwartz2017story}. Consequently, many models lack certain advanced skills such as inference or multiple-sentence reasoning. 

Biases can also come from a few informative key words. For example, entailment models trained on MNLI \citep{bowman2015large} would guess answers based on whether a sentence-question pair contains the same words \citep{mccoy2019right} or solely the existence of keywords \citep{gururangan2018annotation,wang2019if}. Weissenborn demonstrated that more than a third of the questions were answered using a simple baseline model which prioritized answers with question words in the surrounding context \citep{weissenborn2017making}. Sugawara showed that certain questions might require specific lexical patterns around the correct answer \citep{sugawara2018makes}. Researchers have also found certain important words were ignored by MRC models \citep{jia2017adversarial,mudrakarta2018did}, while other less important patterns were overused \citep{mudrakarta2018did}. For example, when negations were added to the questions, datasets such as NewsQA or TriviaQA failed to update their answers \cite{sen2020models}. Other works also found QA models can achieve good performance with incomplete inputs \citep{niven2019probing}.

Furthermore, the questions by themselves sometimes contain clues used by models to locate an answer quickly. As early as 1999, the use of bag-of-words, when combined with other heuristics, achieved up to 40\% accuracy for answering interrogative queries \citep{hirschman1999deep}. Early researchers designed heuristic-rules based systems specifically to answer ‘wh’ questions \citep{riloff2000rule}. In more recent studies, some researchers have found that a notable proportion of the questions were still answerable when incomplete questions were given \citep{sugawara2018makes,kaushik2018much}. Other works showed that the models were not robust when questions were paraphrased \citep{ribeiro2018semantically,gan2019improving}. Chen \textit{et al} also found the existence of spurious correlations in WikiHop which were exploited by the model to achieve good performance using only the questions and answers without the contexts \citep{chen2019understanding}. These studies suggests that keywords in the question allow the model to locate key information without having the model to read and comprehend the context. 

\subsection{Categories of dataset bias in MRC}
Through the literature review, 
we observe that the most commonly seen biases in MRC can fall into three main categories. (1) Some biases directly exploit the relationship between the question and sentences similar to the question (that is, question-sentence pairs with high TFIDF scores), and we refer to them as \textbf{Similarity Bias}. (2) The biases can take advantage of a few key words in context. We refer to them as \textbf{Keyword Bias}. (3) The questions by themselves contain information which can be exploited by models to make predictions without carefully reading the passage. We refer to them as \textbf{Question Bias}.

The three types of biases are closely related to one another. The similarity between the question and the context usually refers to the TFIDF score, which can be understood as the \emph{distance} between them. In fact, each category of bias relies on ‘distance’ at different scales. Similarity bias and keyword bias rely on the sentence-level or the local keyword-level distance from a passage to the targeted question. Likewise, question bias exploits the distance between question tokens and a passage. In fact, this is not a new concept. For example, previous researchers have applied this concept to incorporate distance supervision to enhance their QA models \citep{cheng2020probabilistic}. We are inspired by this abstraction to design our experiments and facilitate our discussion.

\section{Overview of MRCLens}
\label{sec:length}
We are inspired by \citep{sugawara2020assessing} to use ablation experiments to test the impact of biases.
Perturbing the original dataset and reevaluating models using the perturbed data is a method used frequently in various fields of NLP \citep{belinkov2017synthetic, carlini2018audio,glockner2018breaking}. Sugawara and their colleagues presented 12 requisite skills which could be used to evaluate an MRC model. For each skill, they performed one corresponding ablation by perturbing the dataset. A comparison of the performance on the original dataset versus the perturbed dataset would indicate if the specific requisite skill is needed by the model to answer questions. Their method fits the purpose of our study. 
However, the key difference is that,
while they are interested in if specific requisite skills are needed, we aim to study if specific biases are needed by a model.

Our toolkit MRCLens 
incorporates existing works into a new tool which can detect if the biases described above exist in a given dataset at an earlier stage of the training process. MRCLens requires data to be SQuAD formatted and will be provided via github. It consists of three main parts:

(1) A preprocessing module which perturbs the original dataset in 8 ways corresponding to different biases, and tokenizes the data. Specifically, we divide the three categories of biases from section 2.2 into 8 bias units indexed from 1 to 8, and we relate each bias unit to one ablation. Define $\mathbf{X}$ as the feature space, $\mathbf{Y}$ as the labels, $(x, y)$ as an (input, label) pair, and $f$ be a model. Let $b_i$ be a potential bias and $m_i$ be a method which ablates the feature that provides the corresponding information $n_j$. Suppose $f(x)=y$ for some $x$ in $\mathbf{X}$. We are interested in if $f(m_i(x)) = y$, which means $x$ can be solved without information $n_i$.

(2) A neural-network MRC model which trains a model and evaluates it against both the original test data and the perturbed test data. This model is based on a baseline neural-network model put forward by \citep{clark2019don}. After preprocessing, we train a neural-network baseline model on the original training data. Then for each bias, we test the baseline model against the corresponding perturbed dataset. The model’s performance on this new dataset would indicate to what extent the specific bias impacts the result.

(3) An evaluation module which presents the results in an organized format which allows for interpretation. MRCLens compares the performance between the original and the modified dataset. By checking whether the questions are solvable after ablations, we can interpret whether the presence of a specific bias leads to unintended but correct answers. When the performance gap is small, we can infer the bias $b_i$ is used to answer the questions without $n_i$. If the gap is large, a notable proportion of the solved questions may require $n_i$. 

\section{Experiment and Discussion}
\subsection{Experiment Setup}
We use SQuAD (version 1.1) for the experiment. The model is a recurrent co-attention model\citep{clark2019don, chen2016enhanced}. The model consists of an embedding layer with character CNN, a co-attention layer, and a shared BiLSTM layer as the pooling layer. We use a 0.2 dropout rate, a learning rate decay of 0.999 every 100 steps\citep{clark2019don}. We use a plain loss function which computes the negative log likelihood given the model outputs and the labels. Ideally, MRCLens would be agnostic of the model architectures, since we care most about the changes in accuracy before and after ablation, not the accuracy itself.

\subsection{Experiment Results}
We performed four experiments to measure \textbf{Similarity Bias}, which refers to the similarity between a sentence in context and the question calculated based on TFIDF score. In experiments 1 and 2, we inject noise by adding a part of the question or the full question in front of a sentence that does not contain the original answer. This enhances the similarity score between the question and another sentence. If the model relies heavily on the most similar sentence to make predictions, then this change will misguide the model to look for answer span in the wrong place and lead the accuracy to drop. $e_1$ and $e_2$ use a truncated version of the dataset where only one question is kept per passage, because multiple questions are often asked based on one passage but it could be confusing to insert information from all questions.

In $e_3$, we shuffle the sentence order. If the performance doesn't change significantly, that means the model mainly relies on information from individual sentences, but not heavily on the contextual relationship between them. 

\begin{table}[th]
\centering 
\caption{Similarity Bias - f1 drops after $e_1$, $e_2$ and minor change after $e_3$ suggest the model relies on context-question similarity but not so much on the inter-sentence relationships.}
\begin{tabular}{ c c c c } 
\hline
ablation & em & f1 & f1 drop \\
\hline
$e_1$ insert full question &39.72&48.82&30.93 \\ 
$e_2$ insert half question &53.36&64.13&15.62 \\ 
$e_3$ shuffle sentence order&66.19&74.48&6.13 \\
\hline
\end{tabular}
\end{table}

We performed two experiments to evaluate \textbf{Question Bias}. In $e_4$, we keep only the interrogative words in the question, and in $e_5$ we shuffle the order of words in the question. Finally, there are three experiments which measure \textbf{Keyword Bias}. We consider nouns, verbs and adjectives from questions as potential keywords and we insert them respectively to a random sentence in the context other than the one containing the true answer. Like $e_1$ and $e_2$, we use the truncated dev dataset. 

\begin{table}
\centering
\caption{Question Bias - interrogatives alone can still be informative, and the sequence of question words is not essential for making predictions}
\begin{tabular}{ c c c c } 
\hline
ablation & em & f1 & f1 drop \\
\hline
$e_4$ interrogatives words&17.10&23.62&56.99\\ 
$e_5$ shuffle question words &56.08&64.05&16.56\\ 
\hline
\end{tabular}
\end{table}

\begin{table}[th]
\centering
\caption{Keyword Bias - key nouns from questions bring the more noise to contexts than verbs and adjectives.}
\begin{tabular}{ c c c c } 
\hline
ablation & em & f1 & f1 drop \\
\hline
$e_6$ insert key nouns & 51.28 & 62.29 & 17.46\\ 
$e_7$ insert key verbs& 58.68 & 71.07 & 8.68\\ 
$e_8$ insert key adj.& 59.55 & 72.29 & 7.46\\ 
\hline
\end{tabular}
\end{table}
$e_3,e_4,e_5$ use the original dev dataset with 10570 entries whose f1 score is 80.61\%, while $e_1,e_2,e_6,e_7,e_8$ use the truncated dev dataset with 1943 entries and an f1 score of 79.75\%. According to Table 1, accuracies dropped notably due to the added contents from the questions even though everything else remains the same. f1 drops from 80\% to 64.13\% when we insert half of the question, and to 48.82\% when we insert the full question. The model is likely looking for answer in the sentence where question words were inserted, as it is now the most similar sentence. The result from $e_3$ informs us that the sentence order has very little influence on the model's prediction. Thus this dataset is not suitable for evaluating a model's ability to understand `sentence-level compositionality`\citep{sugawara2020assessing}.

Results from Table 3 are consistent with those from Table 1. Our changes shortened the local distance between questions and words or short phrases. The drops in accuracies suggest the models were misled to some extent to search for answers around the inserted words. Nouns retain the most information from questions and thus bring most perturbation to the passages, while verbs and adjectives capture similar amount of information.

Finally, Table 2 suggests the questions alone contain indicative information that could be used when not considered in relation to the passages. In 17\% of the cases, interrogatives are sufficient for the model to make predictions. $e_5$ shows the model's performance is affected only slightly after we shuffle the words to make the question non-sensible.

\subsection{Discussion}
The distances between questions and contexts are indicative of how biased the dataset is. For example, $e_3$ shuffles the sentence order but preserves the distance between sentences and questions, so it has the least effects on the performance. Through experiments 8,7,6,2,1, the noise we inserted to the original dataset gradually lengthens the relative distance between the correct answer. As we add key words or phrases to other parts of the paragraph, the effects of similarity bias or keyword bias are diluted because we enhance the relevance between the questions and other parts of the passages. The drop in f1 score increases from around 8\% to 30.95\% as we increase the noise from inserting keywords to inserting the full questions. 

Our method also provides another way to interpret the similarity bias. The \emph{distance} between the question and the context is one of the most discussed biases in MRC. Indeed, 80\% of our dev dataset has the correct answer in the most similar sentence. $e_2$ inserted the full question into a random sentence in each passage so that the most similar sentence will always be the one where the question was inserted, but despite this change, the model still reached an exact match score of 39.72\%. This suggests the model did not over-rely on the most similar sentence. 

\section{Conclusion}
This study presents a toolkit MRCLens which can be used to detect dataset biases at the early stage of a study. MRCLens can be applied to SQuAD formatted datasets. It outputs helpful interpretations which help researchers to determine to what extent biases exist in the dataset of interest. In future work, we hope to enhance the toolkit to fit datasets of various formats, design methods to quantitatively evaluate the toolkit's outputs, and develop methodologies for other Machine Comprehension Tasks.

%

\bibliography{ref}
\bibliographystyle{icml2022}



\end{document}